\documentclass{article}
\usepackage{spconf,amsmath,graphicx}
\usepackage{amssymb}
\usepackage{cite,multirow}
\newcommand{\R}{\mathbb{R}}
\newcommand{\etal}{\mbox{\emph{et al.\ }}}
\usepackage{amssymb,bm}
\usepackage{graphicx}
\usepackage{float}
\usepackage{cite,multirow}
\pdfoutput=1

\title{Decomposing multispectral face images into diffuse\\ and specular shading and biophysical parameters}

\name{Sarah Alotaibi$^{1,2}$ and William A.~P.~Smith$^1$}
\address{$^1$Department of Computer Science, University of York, UK\\ $^2$Computer Science Department, King Saud University, KSA}
\begin{document}
%
\maketitle
\begin{abstract}
We propose a novel biophysical and dichromatic reflectance model that efficiently characterises spectral skin reflectance. We show how to fit the model to multispectral face images enabling high quality estimation of diffuse and specular shading as well as biophysical parameter maps (melanin and haemoglobin). Our method works from a single image without requiring complex controlled lighting setups yet provides quantitatively accurate reconstructions and qualitatively convincing decomposition and editing.
\end{abstract}
\section{Introduction}

Understanding the appearance of the human face is of considerable interest in computer vision \cite{NeuralFace2017} and graphics \cite{JIMENEZ2010} and has been studied broadly in many works. One approach to understanding appearance is to estimate a physically meaningful decomposition. 
Existing methods for decomposing face appearance often rely on complex equipment and lab conditions. For example, Ma et al.~\cite{ma2007rapid} use polarised spherical gradient illumination to capture geometry and diffuse and specular reflectance maps. Donner et al.~\cite{donner2008layered} use polarised, multispectral light and an assumption of a planar sample to estimate biophysical skin parameter maps. Gitlina et al.~\cite{gitlina2018practical} essentially combine these two setups using multispectral illumination in a lightstage. In a medical setting, Claridge \cite{Claridge2003489} use RGB plus NIR measurements with polarised illumination and again assuming a planar sample to estimate skin biophysical parameters for the purpose of identifying abnormalities.

Highly ambitious recent work uses deep learning to decompose face appearance from single, uncontrolled images. Kim et al.~\cite{Kim_2018_CVPR} and Tewari et al.~\cite{Tewari_2017_ICCV} use self-supervision to learn to fit 3D morphable models \cite{Blanz:1999:MMS:311535.311556} to single images (providing an estimate of geometry, illumination and reflectance). Intrinsic image based methods \cite{NeuralFace2017,sfsnetSengupta18} do not rely on a face-specific statistical model but learn the whole task from scratch.

We consider the challenging problem of decomposing a single image into diffuse and specular shading and the distribution maps of biophysical parameters that explain skin colouration. To make the problem tractable we work with multispectral images but do not require the complex and controlled illumination conditions of previous methods. To do so, we propose a novel combination of a biophysical and dichromatic reflectance model that efficiently characterises spectral skin reflectance. Previously, the dichromatic model has primarily been used in a setting where the light source colour (or more generally spectral power distribution) is unknown but the object under study is assumed to have uniform or piecewise uniform \cite{huynh2010solution} colour (more precisely diffuse albedo or spectral reflectance). We show that, with known light source spectra we can estimate four spatially varying reflectance parameters by fitting our model to multispectral data using nonlinear least squares. We show how the decomposed parameter maps can be edited in an intuitive way and validate our approach quantitatively on a database of skin reflectance spectra and qualitatively on a set of multispectral face images.

\section{Biophysical dichromatic skin model} 

In this section we introduce our multispectral skin reflectance model. We assume that skin reflects light according to the dichromatic model \cite{shafer1985using} and that diffuse albedo (i.e.~subsurface absorption) can be explained using a biophysical model. 

\subsection{Multispectral dichromatic model}

The dichromatic model assumes that wavelength-dependent scene radiance, $L(\lambda)$ with $\lambda$ the wavelength, is a sum of body (diffuse) and surface (specular) reflected components. Further, it divides each source of radiance into a part that depends on geometry (informally ``shading'') and a wavelength dependent part (informally ``colour''). The body reflection arises from subsurface scattering and modifies the spectral power distribution (SPD) of the light through absorption whereas the surface reflectance happens at the interface and does not, meaning the model can be written as:
$L(\lambda) = E(\lambda)(i_dR(\lambda) + i_s)$,
where $i_d\in\R_{\geq 0}$ and $i_s\in\R_{\geq 0}$ are the diffuse and specular shading respectively and $R(\lambda)$ is the spectral reflectance of the diffusely reflected light. We discretise at $D$ evenly spaced wavelengths such that we can write the vector of spectral scene radiance, $\mathbf{l}=[L(\lambda_1),\dots,L(\lambda_D)]^T$, as:
\begin{equation}
    \mathbf{l} = \textrm{diag}(\mathbf{e})(i_d\mathbf{r} + i_s),\label{eqn:discretemodel}
\end{equation}
where $\mathbf{e}\in\R^D$ and $\mathbf{r}\in\R^D$ are the wavelength-discrete illuminant SPD and spectral reflectance respectively.

\subsection{Biophysical skin reflectance model}
We now replace generic spectral reflectance with a biophysical model for skin. This skin model has only two free parameters, meaning the dichromatic model has in total only four unknowns per pixel. Our biophysical spectral reflectance model for skin follows a number of previous models \cite{Claridge2003489,JIMENEZ2010,krishnaswamy2004biophysically,preece2004spectral}, though we focus on simplicity and limiting the number of free parameters. Specifically, our model allows only the melanin and haemoglobin concentration to vary spatially whereas all other parameters are based on measured data, validated approximation functions or average values \cite{jacques1998skin,Alotaibi_2017_ICCV,ANDERSON198113,THODY1991340,JIMENEZ2010,prahl1999optical,Flewelling1999,krishnaswamy2004biophysically}. The free parameters have physical meaning and can therefore be constrained to the range of values observed in healthy skin.

\begin{figure}[h]
	\centering
	\begin{tabular}{@{\hspace{0.01cm}}c@{\hspace{0.1cm}}c@{\hspace{0.01cm}}}
		\includegraphics[height=3.5cm,clip=true]{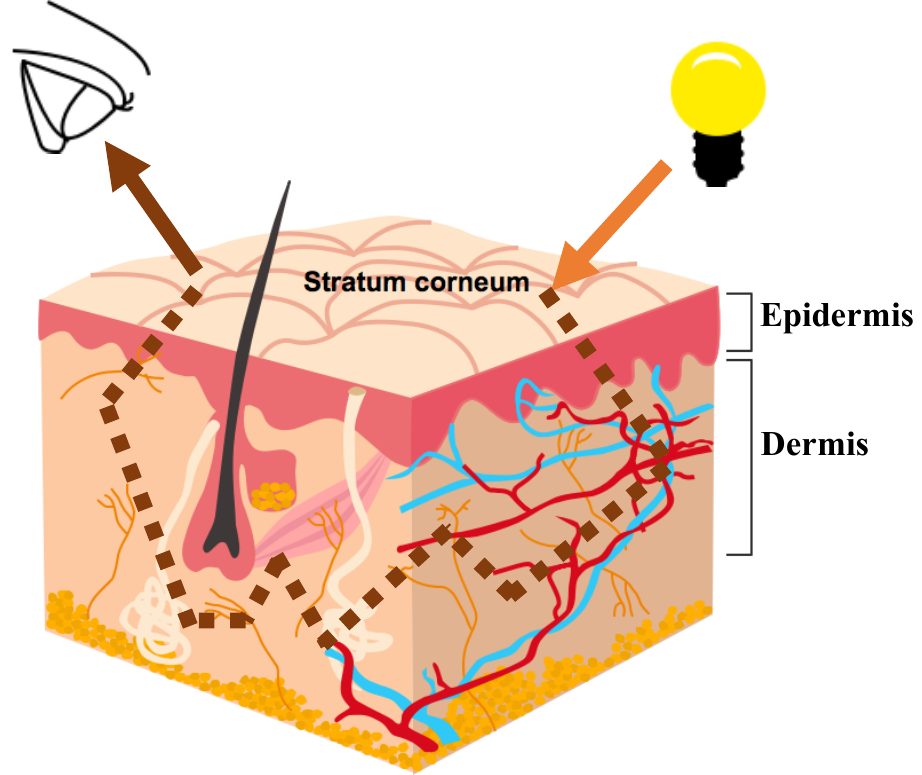}&
		\includegraphics[height=3.5cm,clip=true,trim=0cm 31px 434px 0cm]{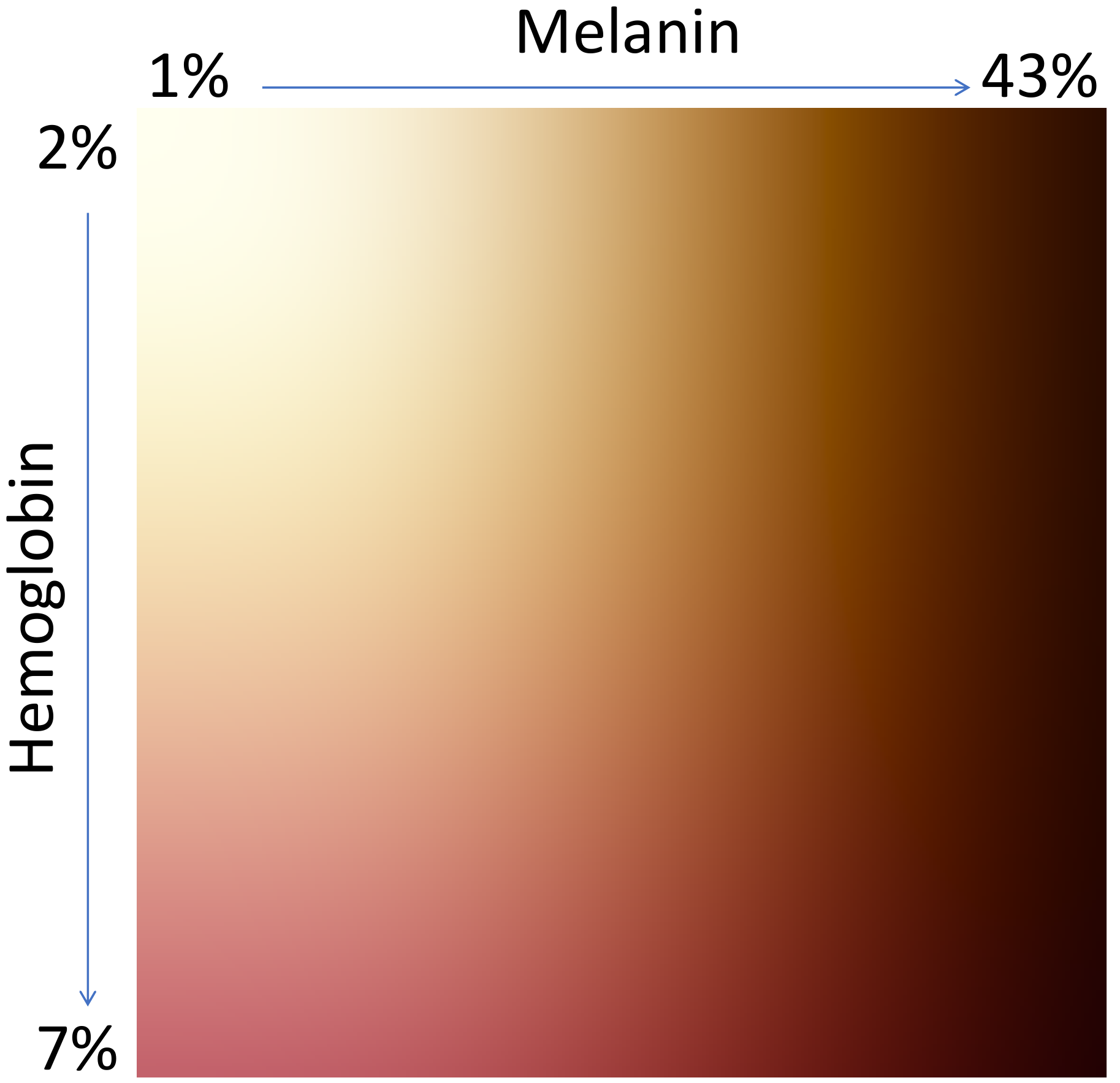}\\
  	    	{\small (a)}&{\small (b)} 
  	\end{tabular}
       \caption{(a) Layered skin reflectance model. (b) sRGB render (see Sec.~4.3) of the skin colouration model.}
 \label{fig:skin_models}
\end{figure}

Human skin tissue has a complicated layered structure. We consider only two layers (Fig.~\ref{fig:skin_models}(a)). The outer layer is the {\it epidermis} contains the melanin and is responsible for absorption of the short wavelengths of the visible spectrum and the remainder of light is mostly forward scattered. The {\it dermis} has blood vessels that contain the haemoglobin pigment, and absorbs light in the green and blue wavelengths, the remainder of light is primarily reflected back through the epidermis where the melanin pigment absorbs the light again. Therefore, our skin spectral reflectance model is written as:
\begin{equation*}
R(f_{\textrm{mel}}, f_{\textrm{hem}},\lambda)= T_{\textrm{epid}}(f_{\textrm{mel}},\lambda) ^{2} R_{\textrm{dermis}}(f_{\textrm{hem}},\lambda).
\end{equation*}
where $f_{\textrm{mel}}$ is the epidermal melanosomes volume fraction and lies in the range $f_{\textrm{mel}}^{\textrm{min}}=1.3\%\dots f_{\textrm{mel}}^{\textrm{max}}=43\%$, $f_{\textrm{blood}}$ is the dermal blood volume fraction and lies in the range $f_{\textrm{blood}}^{\textrm{min}}=2\%\dots f_{\textrm{blood}}^{\textrm{max}}=7\%$, $T_{\textrm{epid}}(f_{\textrm{mel}},\lambda)\in [0,1]$ is the proportion light transmitted through the epidermis (twice) and is modelled using the Lambert-Beer law, $R_{\textrm{dermis}}(f_{\textrm{blood}},\lambda)\in [0,1]$ is the proportion of light reflected from the dermis and is modelled by Kubelka-Munk theory. In Fig.~\ref{fig:skin_models}(b) we visualise the range of colours that can be obtained by our skin reflectance model when transformed to RGB space as described in Sec.~\ref{sec:Spectral2sRGB}. In wavelength-discrete terms once again, we write $\mathbf{r}(f_{\text{mel}},f_{\text{hem}})\in\mathbb{R}^D$ as the vector of diffuse spectral reflectance which can be substituted into \eqref{eqn:discretemodel}.

\section{Model fitting}
\label{sec:Modelfitting}

We now assume that we are provided with a multispectral measurement of scene radiance, $\mathbf{l}_{\text{obs}}\in\mathbb{R}^D$, at a point on the skin surface. We assume that scene illumination is spectrally uniform and that its SPD is known. In this case, our model has four unknown parameters and we pose model fitting as a nonlinear least squares problem whose objective is:
\begin{equation}
    \varepsilon(i_d,i_s,f_{\text{mel}},f_{\text{hem}}) = \left\| \mathbf{l}_{\text{obs}} - \textrm{diag}(\mathbf{e})(i_d\mathbf{r}(f_{\text{mel}},f_{\text{hem}}) + i_s) \right\|^2.
\end{equation}
All four parameters are subject to constraints. The biophysical parameters must lie in their plausible ranges while the diffuse and specular shading must be positive. This leads to a constrained optimisation problem:
\begin{multline*}
    \min_{i_d,i_s,f_{\text{mel}},f_{\text{hem}}} 
    \varepsilon(i_d,i_s,f_{\text{mel}},f_{\text{hem}}),\\ \text{s.t. } i_d\geq 0, i_s\geq 0, f_{\text{mel}} \in [f_{\text{mel}}^{\text{min}},f_{\text{mel}}^{\text{max}}],
    f_{\text{hem}} \in [f_{\text{hem}}^{\text{min}},f_{\text{hem}}^{\text{max}}].
\end{multline*}
In practice, we reparameterise each of the four unknowns to obtain an unconstrained optimisation problem:
\begin{align}
    \min_{\phi_d,\phi_s,\phi_{\text{mel}},\phi_{\text{hem}}}\!\!\!\! &
    \varepsilon\left(i_d(\phi_d),i_s(\phi_s),f_{\text{mel}}(\phi_{\text{mel}}),f_{\text{blood}}(\phi_{\text{blood}})\right),\label{eqn:optprob}\\
    \textrm{where}\ \  i_d(\phi_d) &= \exp(\phi_d),\quad i_s(\phi_s) = \exp(\phi_s),\notag\\
    f_{\text{mel}}(\phi_{\text{mel}}) & = (f_{\text{mel}}^{\text{max}}-f_{\text{mel}}^{\text{min}})S(\phi_{\text{mel}})+f_{\text{mel}}^{\text{min}},\notag\\
    f_{\text{blood}}(\phi_{\text{blood}}) & = (f_{\text{blood}}^{\text{max}}-f_{\text{blood}}^{\text{min}})S(\phi_{\text{blood}})+f_{\text{blood}}^{\text{min}},\notag
\end{align}
and $S$ is the logistic function: $S(x) = 1/(1+\exp(-x))$.
We minimise \eqref{eqn:optprob} using the trust-region-reflective algorithm. Note that, in the case of multispectral images, we solve the optimisation problem independently at each pixel. i.e.~we do not require any spatial smoothness priors.

\begin{figure*}[t!]
\centering
\includegraphics[trim={0cm 0.2cm 0cm 0cm},width=\textwidth,clip=true]{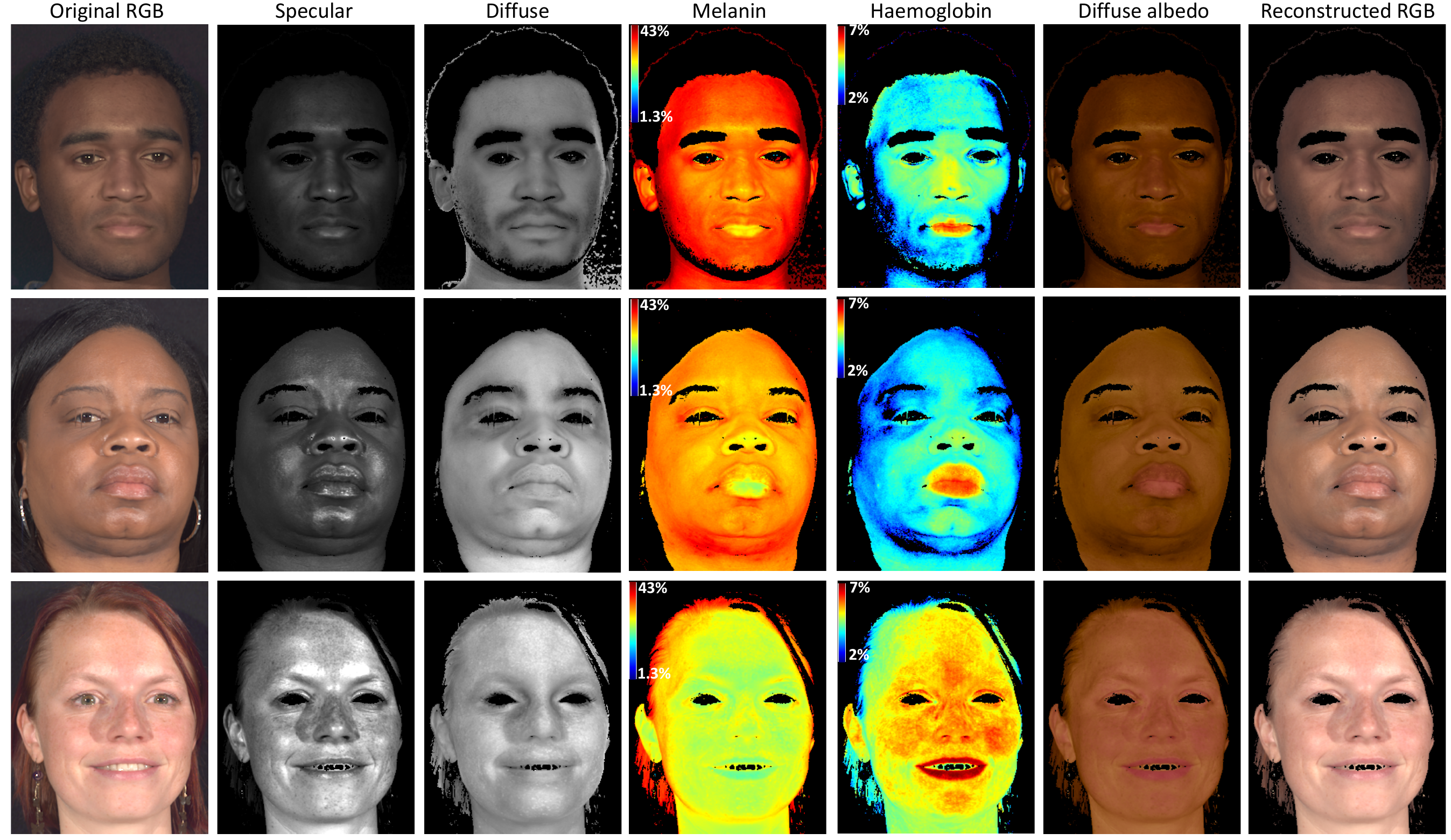}
\caption{Col.~1: sRGB render of multispectral images. Col.~2-5: decomposition into: specular and diffuse shading, melanin and haemoglobin. Col.~6: albedo from biophysical maps. Col.~7: sRGB render of reconstructed radiance.}
\label{fig:Decompostion}
\end{figure*}

\section{Model-based editing}

Once a face image has been decomposed to the four semantically meaningful parameter maps, intuitive editing becomes possible. After editing one or more of the maps, we recompute scene radiance with \eqref{eqn:discretemodel}, blend with unedited background radiance using a skin mask and then transform to sRGB.

\subsection{Learning skin segmentation}
\label{sec:learingSeg}
To produce convincing edited images we require a skin probability mask to blend the edited skin radiance with background.
We train a multilayer perceptron that computes the probability that a single pixel belongs to the skin region. The network has five fully connected layers with ReLU activation (number of channels = 64/64/128/128/2). 
The input for a pixel is a vector $\theta$ comprising the original measured radiance and the four fitted parameters:
\begin{equation*}
\theta =\begin{bmatrix}
{\bf{l}}_{\text{obs}},
f_{\textrm{mel}},
f_{\textrm{hem}},
i_{d},
i_{s}
\end{bmatrix} \in\R^{D+4},
\end{equation*}
The output of final fully connected layer is passed thought the per-pixel classification log loss and the network predicts the probability $p(\text{skin}|\theta)$, for all pixels. 
We train the network by manually selecting skin (total 1.1M pixels) and non-skin (total 750k pixels) regions from the data in \cite{4711794}.


\subsection{Blending skin and background edits}
\label{sec:blend}
Our biophysical model is able to describe only skin spectral reflectance. So, other features such as facial features: eyes, teeth, facial hair and the image background are not well explained. For this reason, we use the estimated skin probabilities to blend the spectral radiance for non-skin regions and our edited biophysical spectral radiance for skin regions:
\begin{equation*}
   \mathbf{l}_{\text{edit}} = p(\text{skin}|\theta)(i_d\mathbf{r}(f_{\text{mel}},f_{\text{hem}}) + i_s) + (1- p(\text{skin}|\theta)) \mathbf{l}_{\text{obs}}.
\end{equation*}

\subsection{Spectral radiance to nonlinear sRGB}
\label{sec:Spectral2sRGB}
For visualisation, a spectral radiance vector $\mathbf{l}_{\text{edit}}$ resulting from an edit must be rendered to a tristimulus RGB image. This arises from an integration over wavelength of the product of scene radiance (itself the product of illumination and reflectance spectra) and camera spectral sensitivity:
\begin{equation}
    i_c = \int_0^{\infty} E(\lambda)R(\lambda)S_c(\lambda)d\lambda,
\end{equation}
where $E$ is the SPD of the illuminant, $R$ the spectral reflectance of the surface and $S_c$ the spectral sensitivity of the camera in colour channel $c\in\{R,G,B\}$. In our wavelength-discrete formulation this becomes:
\begin{equation}
    \mathbf{i}_{\textrm{sRGB}} = [i_R, i_G, i_B]^T = \left(\mathbf{T}(\mathbf{S},\mathbf{e})\mathbf{S}^T\textrm{diag}(\mathbf{e})\mathbf{l}\right)^{1/\gamma},
\end{equation}
where $\mathbf{S}\in\R^{D\times 3}$ contains the wavelength-discrete versions spectral sensitivities of the camera, $\mathbf{T}(\mathbf{S},\mathbf{e})\in\mathbb{R}^{3\times 3}$ is a matrix that performs white balancing and colour transformation to sRGB space and $\gamma=2.4$ controls the nonlinear gamma.

\section{Experiments}


\begin{figure}[t!]
\centering
\includegraphics[trim={0cm 1cm 0cm 0cm},width=\linewidth]{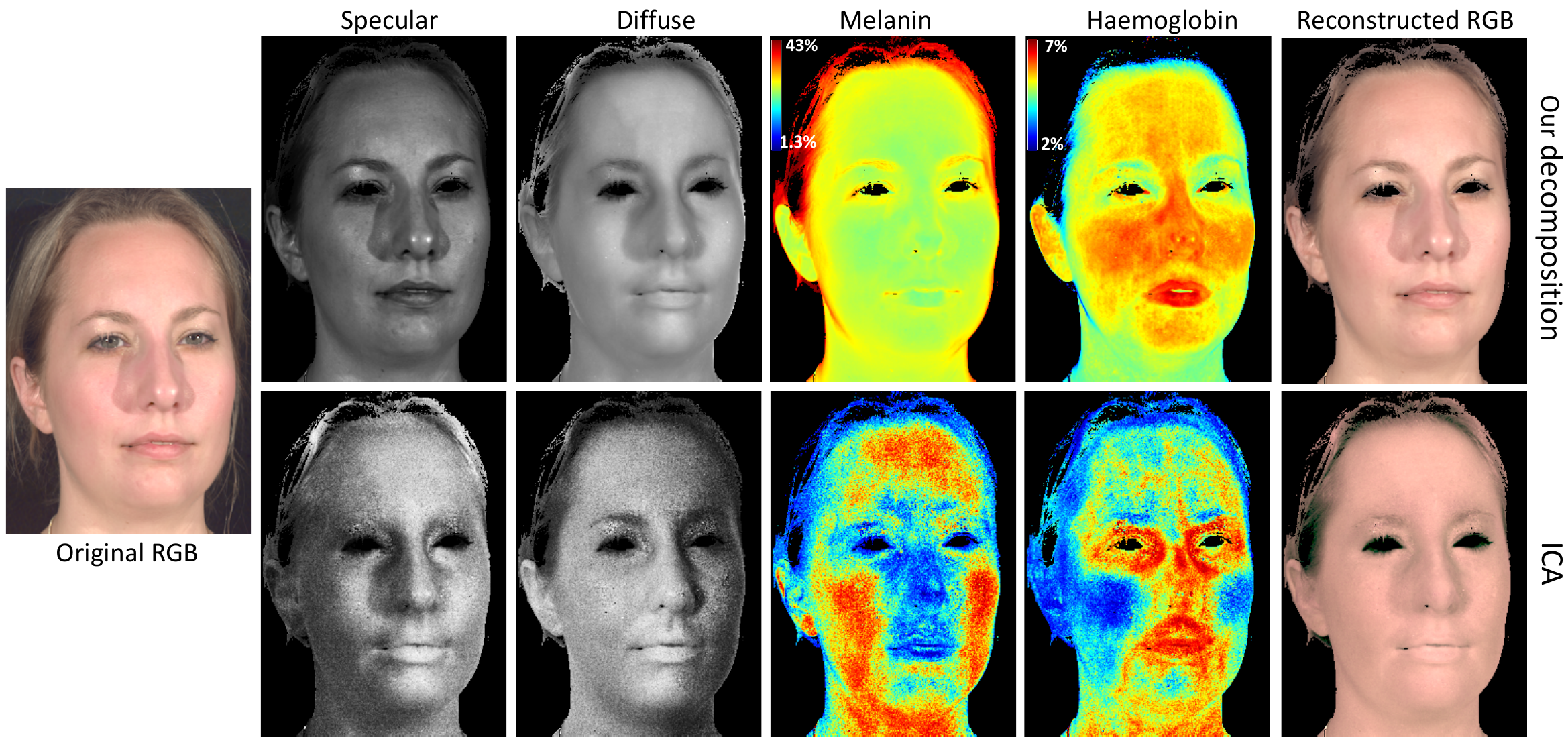}
\caption{Qualitative comparison to \cite{Tsumura:2003:ISC}, from left to right: sRGB render of multispectral image, specular shading, diffuse shading, melanin map, haemoglobin map and the reconstructed sRGB. The top row is our result, bottom is \cite{Tsumura:2003:ISC}.}
\label{fig:ICAexp2}
\end{figure}

We begin by quantitatively evaluating how well our model and model fitting method are able to explain measured data. We use the 25 faces from the ISET database \cite{4711794}, compute the best fitting parameters for skin labelled pixels and then compute the mean relative absolute error in the reflectance spectra over all pixels in all images, yielding a 7.4\% error.

In Fig.~\ref{fig:Decompostion} we show representative qualitative results from this experiment. Input, albedo and reconstruction RGB renderings use the mean camera spectral spectral sensitivity from \cite{6475015} and assume daylight (D65) illumination. The specular shading maps clearly pick up the specular reflections. Note also that the specular shading contains fine surface detail whereas the diffuse shading is blurred due to subsurface scattering (consistent with \cite{ma2007rapid}). The diffuse albedo is computed directly from the biophysical spectral reflectance and shows that shading effects are not transferred to the biophysical parameters. Note that lips and flushed cheeks are clearly visible in the haemoglobin maps and that the overall melanin value accurately reflects skin colour. All results are masked to skin region by binarising the skin probability maps.

There are no existing methods solving the same task, but for comparison we adapt the method of Tsumura \etal \cite{Tsumura:2003:ISC} to multispectral data. This is a statistical approach based on Independent Component Analysis (ICA) so the parameter maps have no exact physical meaning. However, as seen in Fig.~\ref{fig:ICAexp2} the maps do seem to have an approximate correspondence to our physically meanginful ones, though it is clear that our decomposition and reconstruction is more plausible.

\begin{figure}[!t]
\centering
\includegraphics[trim={0cm 0.5cm 0cm 0cm},width=\linewidth]{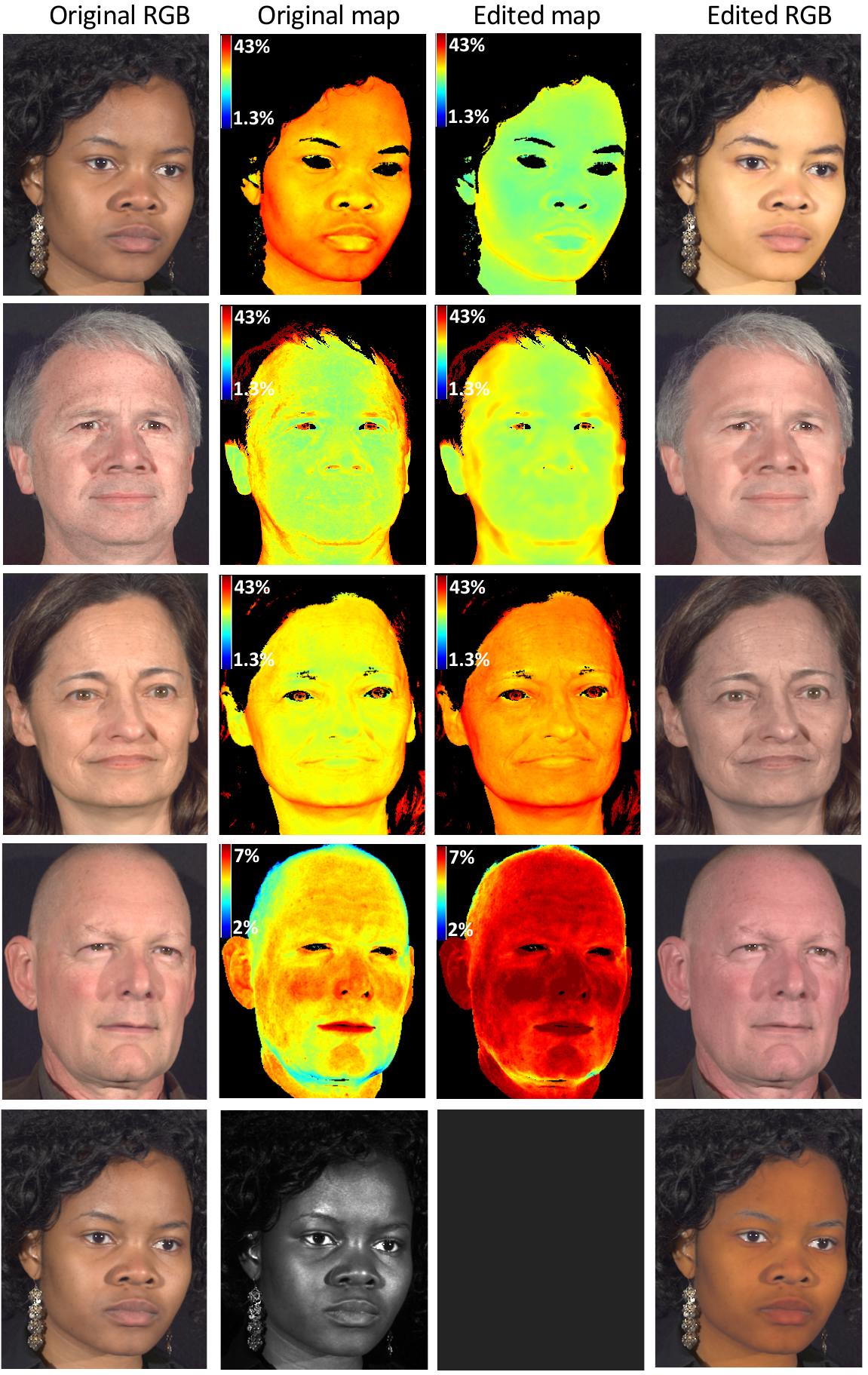}
\caption{Editing the decomposed maps. Rows 1-3 show the results of editing melanin map. Row 4 presents the result of increasing the haemoglobin. Last row shows an application of specularity removal.}
\label{fig:BioEditing}
\end{figure}


Finally, in Fig.~\ref{fig:BioEditing} we present editing results using the method in Sec.~4. In each case we show the original RGB and one of the estimated maps, then the edited map and finally the edited RGB result. In the first row we scale the melanin by 0.75, giving the appearance of lighter skin. In the second row, we show freckle removal by applying 2D median filter on the melanin map. In the third row we increase the melanin concentration by 0.2 and this shows the face appearance with darker skin as if the face were sun-tanned. In the fourth row we increase the haemoglobin map by 0.3. This presents a flushed appearance of the face such as if the face is over heated. Finally, in the fifth row we demonstrate specularity removal by setting the specular map to a constant.

\section{Conclusion}

We have presented a novel hybrid of a spectral biophysical and dichromatic reflectance model and shown how to fit the model to multispectral face images. Our results show that the model is able to accurately explain multispectral skin reflectance and that the estimated maps provide a highly plausible decomposition. The surprising conclusion of our work is that there is sufficient information in a multispectral image to render the decomposition task well-posed. This is possible only by introducing the constraint of a biophysical model. In principle, our model could be fitted to four channel data so it would be interesting to see whether we can obtain similar results using RGB + NIR images. Our inverse rendered results could be used to learn statistical models of the variation in intrinsic biophysical face parameters \cite{Alotaibi_2017_ICCV}.

\bibliographystyle{IEEEbib}
\bibliography{refs}
\end{document}